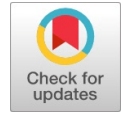

# Stereo Vision Based Robot for Remote Monitoring with VR Support

**Mohamed Fazil M. S., Arockia Selvakumar A., Daniel Schilberg**

*Abstract: The machine vision systems have been playing a significant role in visual monitoring systems. With the help of stereovision and machine learning, it will be able to mimic human-like visual system and behaviour towards the environment. In this paper, we present a stereo vision based 3-DOF robot which will be used to monitor places from remote using cloud server and internet devices. The 3-DOF robot will transmit human-like head movements, i.e., yaw, pitch, roll and produce 3D stereoscopic video and stream it in Real-time. This video stream is sent to the user through any generic internet devices with VR box support, i.e., smartphones giving the user a First-person real-time 3D experience and transfers the head motion of the user to the robot also in Real-time. The robot will also be able to track moving objects and faces as a target using deep neural networks which enables it to be a standalone monitoring robot. The user will be able to choose specific subjects to monitor in a space. The stereovision enables us to track the depth information of different objects detected and will be used to track human interest objects with its distances and sent to the cloud. A full working prototype is developed which showcases the capabilities of a monitoring system based on stereo vision, robotics, and machine learning.*

## I. INTRODUCTION

Real-time camera monitoring from remote's applications has been grown in recent time because of its ability to continually keep monitor any place from anywhere in this world. However, it lacks a realistic view as it produces only 2D images with no depth information and not natural motion control. PTZ (Pan, Tilt, and Zoom) cameras [4] are capable of covering a wide area with its movement, but it still lacks flexibility where the user cannot naturally view things like he would experience when he is really at that location. This project's motive is to provide the user with a realistic real-time experience with the emergence of Virtual Reality accessories available for various –internet devices, i.e., Google Cardboard based goggles (e.g., VR Box Fig.3). In it, a 3D graphics environment is projected in a stereoscopic format where the user will experience it by moving the head naturally [8].

Here instead a real-time video stream of calibrated stereo video (Fig.2) from the camera can be used to stimulate the same experience with the help of image processing, robotics, and the internet of things.

Use of deep learning can make this robot more intelligent and environment-aware (Fig.2). It provides us with a base for HRI (Human-Robot Interaction) [6] [1] where it senses and interacts with its environment or transfers one's interaction remotely. This robot uses deep neural networks to find and track objects and faces like a human and stream it to the user where the user will able to monitor things effortlessly with 3D experience using VR [9] from remote (Fig.3).

Stereovision is a cost-effective way of sensing depth information from the environment as compared with ROS Kinect which is popularly used technology in depth sensing. ROS Kinect [9] works as an active 3D depth estimation system that uses IR laser structured patterns for Depth estimation and is reliable only in indoor applications. Stereo Camera system (Fig.1) is used to compute projection matrix that will transform a 2D point into a 3D point in the left camera coordinate system [2]. These stereo camera systems are way more reliable in outdoor applications compared with Kinect. By sensing the depth, the robot will be capable of maintaining the target object in 3d-space.

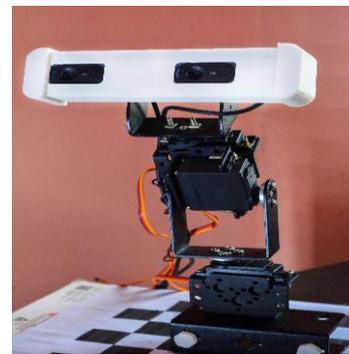

**Fig. 1. 3-DOF Stereovision robot**

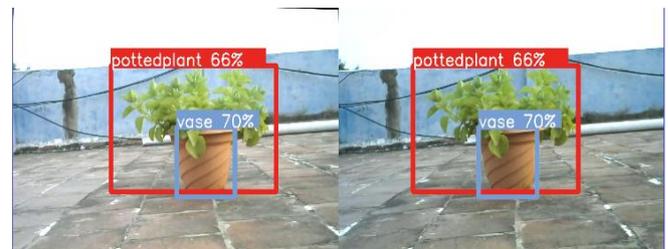

**Fig. 2. Undistorted Stereoscopic Video Feed from stereo camera with Deep Learning used to detect objects with its probability percent**

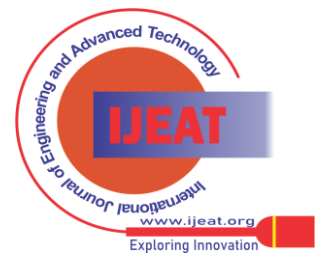

# Stereo Vision Based Robot for Remote Monitoring with VR Support

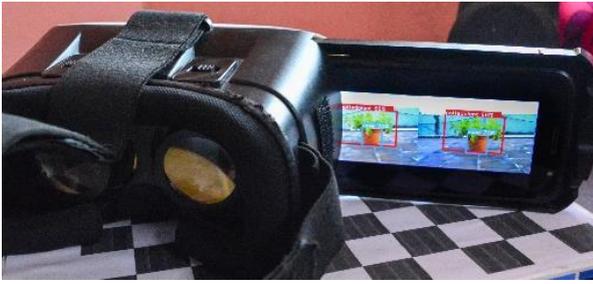

**Fig. 3. Stereo video feed streamed to a smartphone in VR Box through the web server**

## II. SYSTEM DESIGN OVERVIEW

This robotic system is designed to work as a human natural head movement-controlled vision system and also as an autonomous object tracking robot. The user will be able to access the robot through the cloud interface created to choose a movement-controlled or specific object or person tracking with the help of deep neural networks and trained data sets. Hence, this robot provides enough flexibility for the user to utilize all its capabilities.

This system is made up of three levels. The first level is the standalone robot (Fig.4) system which consists of the stereo camera coupled to the 3-DOF robot manipulator as the end effector (Fig.1). In this setup the stereo cameras are connected to a windows laptop with Intel i5 2.4 GHz processor for experimental purpose. The image processing and deep learning are done on the computer. The 3-DOF robot consists of three MG955 servos each with 11kg-cm max torque and capable of doing 0.2 sec/60o. The servo controls are given through an Arduino Uno which will act as the controller (Fig.4) which will take commands through the serial com port. The images processed stereo video feed will be sent to the cloud server and will also receive user inputs object for autonomous tracking or real-time control with the user's head movement control. The next level is the cloud server (Fig.4) which will act as the interface between the standalone robotic system and the internet devices connected to it. In this setup, the web server would provide an interface for the internet devices to interact with the robot. The third level is the end user's remote internet device (Fig.4) which will be provided an application to receive the video stream and return inputs to the cloud server.

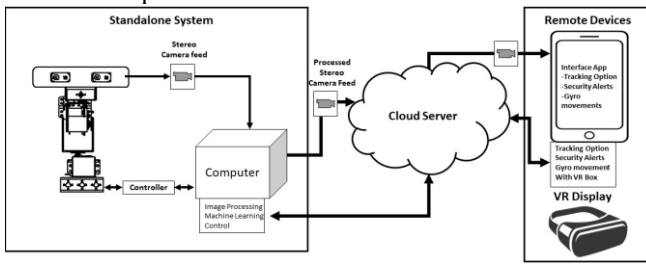

**Fig. 4. Proposed Stereovision Based monitoring robot system**

## III. PREVIOUS RELATED WORK

Lei Chen [1] developed a stereovision only based interactive mobile robot for HRI application. It uses RGBD sensors designed for this application on a Pioneer 3-AT all-terrain robot platform. Their robot has three-level architecture, i.e. sensors, processing, and perception which was able to identify humans of interest and maintain face to face interaction. Xiai Chen [5] presented a 2-DOF robot with stereo cameras which was enabled with face recognition using adaboost face recognition system and 2DPCA face recognition algorithm. It was able to track faces, and locking function was implemented. Mehdi Ahmadi [3] presented a new method for real-time video streaming system for remotely monitoring ultrasound exams which allow streaming ultrasound images including the probe position with latency less than 1 second for HD images. In this paper, we went a step ahead in which we incorporated deep neural networks to detect and classify objects combined with face recognition algorithm. The 3-DOF robot to mimic human head capable movements. It also presents a method to stream the stereo vision to the cloud for real-time remote monitoring solution with VR support.

## IV. DESIGN OF 3-DOF ROBOT

This 3-DOF robot is designed with keeping the human head's motion in mind [6]. A human head is capable of 3-DOF simultaneously are yaw, pitch, and roll. Moreover, the stereovision camera which will mimic human eyes is placed at the end effector of this 3-DOF robot (Fig.5). The robot will be able to achieve 900 of rotation from its neutral position in both directions, i.e., clockwise and anti-Clockwise with each servo motor. This will enable the robot to achieve realistically human head like motion and placing its eye, i.e., stereo cameras in the spatial frame of actual human eyes in a head are capable of (Fig.6).

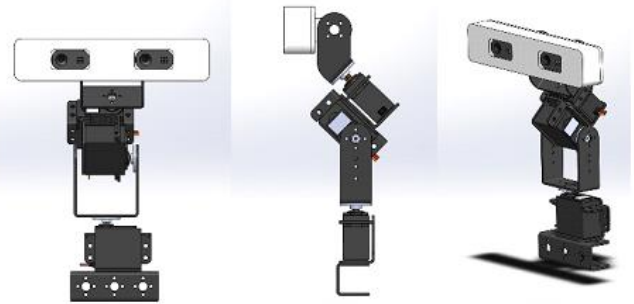

**Fig. 5. 3-DOF Robot manipulator with the stereo camera as its end effector**

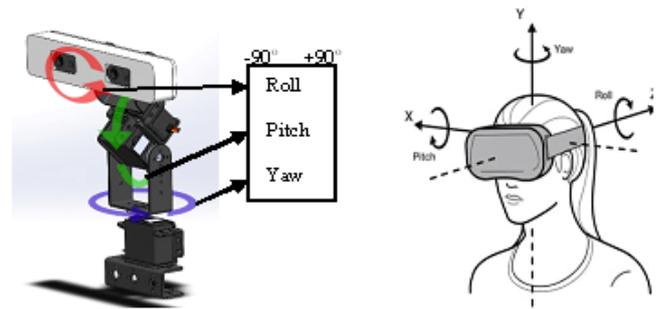

**Fig. 6. Yaw-Pitch-Roll motion comparison with a human head with VR**

The gyroscope's yaw, pitch, and roll (in degrees) (Fig.6) sensor data of the smartphone will be streamed to the standalone system when in VR control mode through the cloud server. This yaw, pitch and roll data in degrees is used to position the corresponding servo motors.

## V. STEREO VISION SYSTEM

Humans are very good at interpreting the depth information from the environment because they have two eyes. With the help of stereo correspondence, i.e., where we capture two images of the same scene from different viewpoints (average distance between our two eyes in 62mm) where our brain perceives a 3D map using the differences in the scene from both the viewpoints images.

The dual cameras present in a stereo camera setup are supposed to be highly similar and have cohesive optics between the both, but commercially available pinhole cameras do not have that closeness due to manufacturing minute differences in the lenses which causes distortions [6] mostly radial distortions and some tangential distortions. For this setup, a regular two Logitech C270 HD webcams (Fig.7) are used which has a max 720P resolution. For image processing, purpose 640X480 resolution was used.

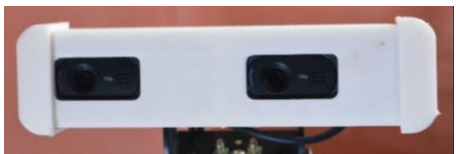

**Fig. 7. Stereo Camera setup**

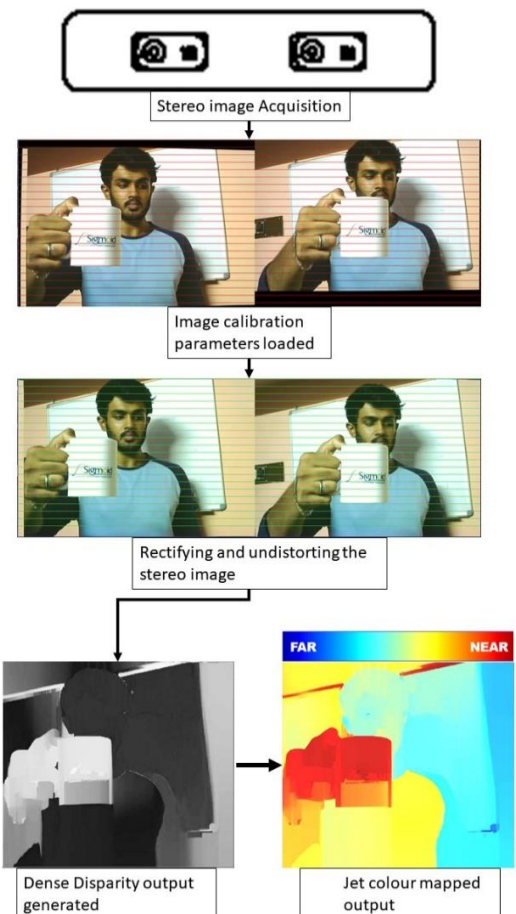

**Fig. 8. Stereo-Cameras depth mapping process.**
(a) Lighter the intensity level is closer than the object is, and the black areas represent unconfident areas.
(b). The grayscale disparity map is colour mapped to Jet for better visual understanding

Both the cameras are rigidly mounted with 72mm distance between them using a 3d printed case. As mentioned, these both webcams would also produce distortion within each other. Therefore, it is required to calibrate and undistorted the images (Fig.8) coming from both the cameras to give us consistent with both the optical axis parallel with each.

Using the chessboard camera calibration method both the calibration model for both the cameras is generated which can be used further in any program to undistorted and rectify the stereo images. The depth information can be gained from the stereo images by using stereo correspondence and generating dense disparity image which represents the depth information (Fig.8). All the image processing is done using the OpenCV library in the python environment.

## VI. HARDWARE DESIGN OF THE 3-DOF ROBOT

The 3-DOF robot consists of three MG995 servo motors controlled by an Arduino Uno which will be receiving serial commands for the movement of the motors. The hardware design specifications are given in Table1. The servo motors will be controlled using PWM signals generated by the Arduino. For the cameras two Logitech C270 webcams were used. The computer which was connected to the Arduino using USB serial interface will send the calculated next position values (Pitch, Yaw, Roll) from the python program used for image processing. To reduce the cost and increase flexibility of the design, regular available standard aluminium servo brackets were used to build the robot body.

**Table 1. Hardware Specification**

| Control Board | Arduino Uno Rev3 (AT mega 328p, 16 MHz, 32 KB flash memory, USB, I2C, 6-20V) |
|---|---|
| Servo Motor | TowerPro MG995, 10kg-cm @ 4.8V |
| Power Supply | Low Cost 2 Cell Li-ion batteries (7.4 V) |
| Test Computer | Windows 10 PC, Intel i5-6th Gen, 16 GB Ram, Nvidia GTX 1060 6 GB |
| Cameras | Logitech C270, Video Resolution – 720 x 1280, Field of View – 60º, 1.2 mp |

The position values calculated by the python program with the help of image processing send it to the serial interface with no delay. The Arduino handles the received data with delay of 20 milliseconds for the servo to function smoothly. This is how the synchronisation of the head movement with the robot is also achieved in Realtime where the actual position values of the connected mobile device are wirelessly transmitted to the computer using HTTP and the computer sends it to the controller as a serial data (Fig.10).

## VII. OBJECT DETECTION AND FACE RECOGNITION USINF DEEP NEURAL NETWORK

For object detection-classification, Mobile Net-SSD Network deep learning module is used. SSD, i.e., Single shot multi-box detector are popular and more versatile object detection algorithm.

## Stereo Vision Based Robot for Remote Monitoring with VR Support

A regular CNN (Convolutional Neural Network) gradually decreases the feature map size and increases the depth as it goes deeper in layers. The deep layers cover large receptive fields and construct more abstract representation whereas the shallow layers cover the smaller receptive fields. By using this technique and information from it, the shallow layers are used to predict small objects, and deeper layers are used to predict larger objects because small objects do not require more significant receptive fields.

MobileNet-SSD uses Single Shot Multibox Detector algorithm where MobileNet is feature extractor. MobileNet-SSD generally gives more frame rate per second compared with Fast R-CNN. YOLO is the first framework to reach real-time detection standard with 45 FPS (on GPU) and a mAP of 63.4% on VOC2007, but still has a drawback in detecting smaller objects. This was later remedied by SSD [12] by combining the anchor box proposal system of faster-RCNN and using multi-scale features to do detection layer. SSD further improved mAP on VOC2007 to 73.9% while maintaining similar speed as YOLO [11].

Even though YOLO gives more frames per second, it lacks accuracy. Moreover, Mobile-SSD is also less hardware intensive (can be used in single board computers) compared with other deep learning module but gives fewer proposals. MobileNet-SSD is the most optimal deep learning module with a balanced tradeoff of speed, accuracy and hardware intensiveness as mentioned in Table 2. For this setup, a MobileNet pre-trained model that was trained in the Caffe-SSD framework is used. This model can detect 20 classes.

**Table 2. Speed(ms) vs. Accuracy (mAP) tested on MS COCO cited in [14]**

| Model | mAP | GPU Time (ms) |
|---|---|---|
| SSD321 | 28 | 61 |
| YOLOv2 | 21.6 | 25 |
| SSD513 | 31.2 | 156 |
| R-FCN | 29.9 | 85 |

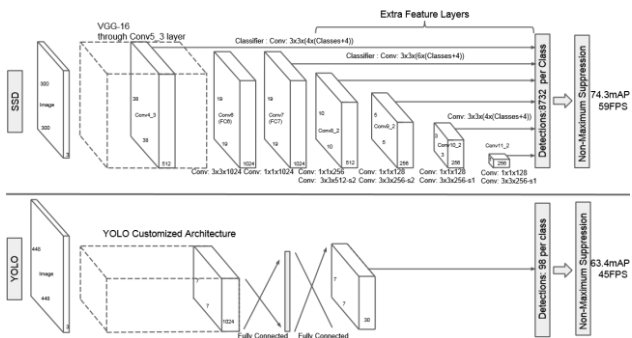

**Fig. 9. A comparison between two single shot detection models:**

SSD and YOLO. The SSD model adds several feature layers to the end of a base network, which predict the offsets to default boxes of different scales and aspect ratios and their associated. Cited in [12]

For facial recognition, FaceNet [10] deep learning module was used. FaceNet is a deep learning architecture which consists of convolutional layers based on the Googlenet inception model. FaceNet is also a one-shot detector model that directly from face images. FaceNet uses TensorFlow which is a deep learning platform in python. Using the trained model different person names with their images (30 to 50 pictures) can be trained into classes for facial recognition. A classifier file will be generated which will contain the face match with the label data which will be used to recognize different faces.

### VIII. CLOUD SERVER AND INTERNET DEVICES

The data exchange between the cloud server and the remote client device are doing with the help of the HTTP protocol which is intended to be quick and lightweight [3]. There are two parts to this system that is the host computer running the Apache web server and the client side which is a smartphone (Fig.11). For this setup, an Android smartphone is used. At the host computer, an HTML webpage is used to give HTTP requests and posts to the client device. The host computer which is a standalone system with the robot will have a python program running as an interface between the server and the robot (Fig.11). In the host computer, the stereo video feed received after all the image processing mentioned above (Fig.8) is converted into Motion JPEG images and it is streamed to the server as chunks of files via Flask web app which is a web management library. The client device can now access the stereo video feed through the video feed IP address of host machine and port in the LAN or WAN if port forwarded. A latency of below 100 milliseconds can be achieved in a Realtime scenario during streaming.

The host computer will be requesting two data from the client device. The first one is the Yaw, Pitch and Roll values from the gyroscope sensor in the smartphone which will be used to control the robot manually with VR head motion. The other data is the subject of interest to monitor when VR head motion is disabled, and the robot starts to track the subject given by the user.

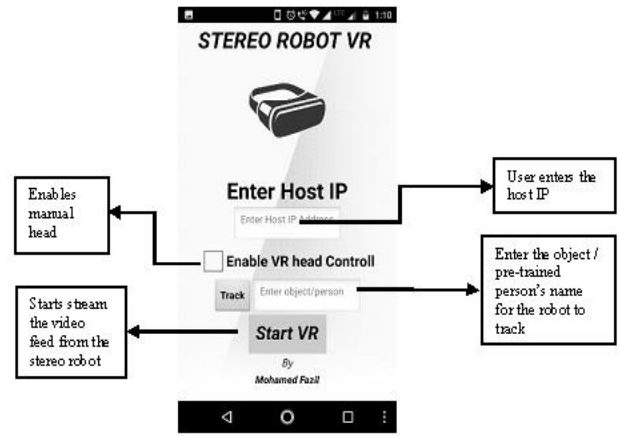

**Fig. 10. Android Application created to interface smartphone with the cloud server**

An android application (Fig.10) was created using the MIT App Inventor which is a web-based Android app development environment. The app will take the input address of the host computer and get it connected to the cloud server.

They provide other options for the user, i.e., enabling the manual VR head control which will make the robot move with the user's head motion by sending the gyro sensor values, and the other option is to make the robot autonomously track the user's subject of interest. This app will post request to get the video feed as response and streams it in VR viewable mode (Fig.3). The app can also be used to do a digital zoom of the acquired video transmission till 1.6 times for the test model camera due to sharpness limitation. But digital zoom can increase even more with better image resolution and sensors bigger than micro thirds.

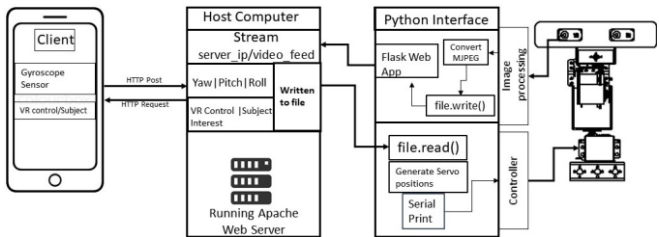

**Fig. 11. Cloud Server – Client interaction and schematics**

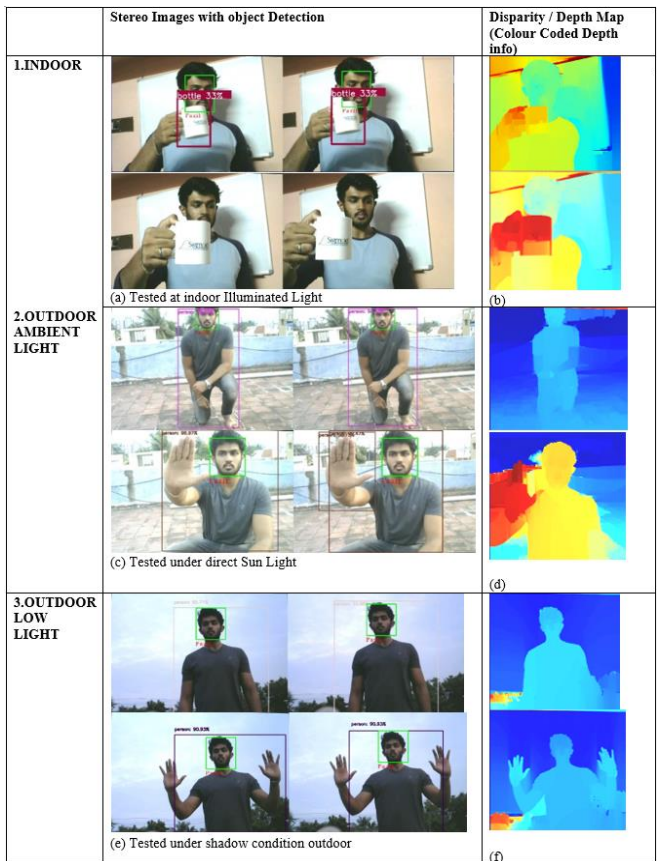

**Fig. 12. Stereo Camera Depth Map Results**

## IX. RESULTS AND DISCUSSION

The robot was tested in different environment such as indoor, outdoor and in different lighting conditions to test its versatility in handling object tracking with deep learning, depth perception and streaming the stereo video feed to the cloud server connected with the remote device. The indoor test (Fig.12 a,b) results were good, and the face recognition, object detection, and depth mapping were precise. In outdoor ambient light and low light, tests were done. In ambient lighting (Fig.12 c,d) the stereo camera was able to depth map till the wall behind (Fig.12 c) which was 17 feet away from the cameras. In low light test (Fig.12 e,f) the face recognition had some inaccuracies, but object detection and classification worked well. In low light, too depth mapping was accurate enough to distinguish objects in 3d space but wasn't accurate enough to find distances between objects. From the test case results, it can be inferred that proper depth perception is being less accurate when it comes to finding distances. However, this can be solved by using night vision cameras with proper calibration to get a perfect cohesive image pair. However, for distinguishing objects in 3D space, this result is mostly enough. The object detection in 3D space can also be improvised by using machine learning to detect probable patterns of error and develop reliable object detection and classification systems.

## X. SUMMARY AND FUTURE SCOPE

In this paper, we have presented a new robotic system for remote visual monitoring that mimics a human head function with the help of stereovision, deep learning, internet devices and a 3-DOF robot to mimic the movements of a human head. This project provides a realistic visual monitoring system apart from the regular TPZ cameras available. With the help of stereo cameras and image processing, we were able to achieve a quality real-time 3D experience in VR by just using two webcams for stereovision and a 3-DOF robot manipulator where any mobile device with a low-cost VR box will be able to experience it which is highly cost effective. The deep learning permits us to train different models combining object detection, facial recognition, and depth perception information to make advanced predictions and perception about the environment. The homemade 3D printed stereovision system after proper calibration was able to depth map till 17 feet in outdoors which proves to be a reliable, low-cost depth sensing technique. An android application was created to interface with the cloud using HTTP in LAN. A latency of about 100-150 milliseconds was achieved with deep learning and video transmission to the VR Set up. The user-friendly android application for VR view is a very versatile soft were it receives the stereo images feed at the same time sends the user head position in real time to the robot with low latency as 60 milliseconds to the robot. This paper is a concept proof how an IOT system with computer vision can be used for advanced and realistic monitoring solutions in homes and industries. In future, even more, fail proof MQTT protocol which is leading in IOT systems can be used for active peer to peer communication.

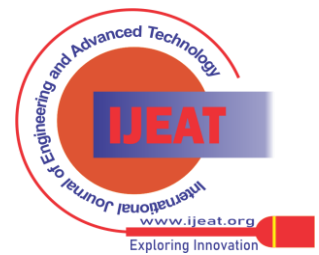

# Stereo Vision Based Robot for Remote Monitoring with VR Support

## AUTHORS PROFILE

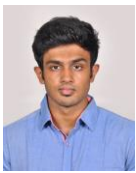

**Mohamed Fazil M. S.**, Currently Pursuing Master's in Mechatronics and Robotics at Tandon School of Engineering, New York University. Have been doing projects and research work on assistive technology, computer vision robotics and IOT. Have received the 'Best Project Award' for two consecutive academic years at Vellore Institute of Technology, Chennai.

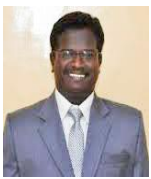

**Dr. Arockia Selvakumar A,** is member of Design and Automation Research Group at the Vellore Institute of Technology, Chennai where he has been since 2013 as an Associate Professor of School of Mechanical and Building Sciences. His field of interest includes Robotics and Automation, IOT, Bio-Mechanics and Finite Element analysis.

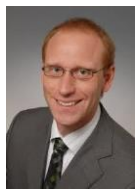

**Prof. Dr. Daniel Schilberg,** is a Vice Dean Mechatronics and Research and Head of Institute of Robotics and Mechatronics University of Applied Sciences Bochum.. Germany. His field of interest includes Robotics, Mechatronics and Digital Transformation.